\documentclass[conference,a4paper]{IEEEtran}
\pdfoutput=1
\ifCLASSINFOpdf
 
\else
  
\fi

\usepackage[english]{babel}
\usepackage[breaklinks=true]{hyperref}
\usepackage{url}
\usepackage{nameref}
\usepackage{amsthm}
\usepackage{amsmath}
\usepackage{graphicx}
\usepackage{color}
\usepackage{nccmath}
\usepackage[overload]{empheq}
\usepackage{hyperref}
\DeclareGraphicsExtensions{.pdf,.png,.jpg}

\DeclareRobustCommand*{\IEEEauthorrefmark}[1]{\raisebox{0pt}[0pt][0pt]{\textsuperscript{\footnotesize #1}}}

\begin{document}
%
\title{Using Scratch to Teach Undergraduate Students' Skills on Artificial Intelligence}

\author{\IEEEauthorblockN{
Julian Estevez,  
Gorka Garate,   
JM Lopez-Guede,   
Manuel Gra\~na      
}                                     
\IEEEauthorblockA{ 
 University of the Basque Country (UPV-EHU), Spain, julian.estevez@ehu.eus}

}

\maketitle

\begin{abstract}
This paper presents a educational workshop in Scratch that is proposed for the active participation of undergraduate students in contexts of Artificial Intelligence. The main objective of the activity is to demystify the complexity of Artificial Intelligence and its algorithms. For this purpose, students must realize simple exercises of clustering and two neural networks, in Scratch. The detailed methodology to get that is presented in the article.
\end{abstract}

{\keywords Artificial Intelligence, Scratch, Neural Network}

\IEEEpeerreviewmaketitle

\vspace{7pt}
\section{Introduction}\label{sec:Intro}
Scientific Method is key in the development of technologically advanced communities \cite{OECD}, and it is of great importance that our students include it in their curricula. As many other competences, it is convenient that some knowledge of scientific  method is learned at the undergraduate stage, and any effort done in the Educational System to foster the learning of it undoubtedly pays.

For a better understanding of scientific method, the Educational Community widely recognizes that school curricula must move on from traditional expositive classes to more informal and collaborative contexts. The fact is, however, that active participation of the students is difficult to achieve at the classroom. One of the reasons should appeal to the interests of the students themselves; in order to achieve active involvement, the classes should make use of more attractive resources, such as fun, games, social interaction, observation of real problems, novelty, etc. This is specially clear in the case of undergraduate students. Surprisingly, there is a quite widespread mistrust of science in the post-truth era where we live. This makes it necessary to insist on a bigger effort to spread the benefits of science among younger students \cite{Arimoto},\cite{Haerlin}.

The present article presents the design and development of several simple educational exercises to promote understanding and learning of Artificial Intelligence (AI) at schools with the usage of Scratch, which is a graphic programming environment. By allowing novices to build programs by snapping together graphical blocks that control the actions of different actors and algorithms make programming fairly easy for beginners.

AI is one of the technologies that will transform our society, economy and jobs in a greater extent along the next decades. Some of the most known examples of AI are driverless cars, chatbots, voice assistants, internet search engines, robot traders, etc. These systems can be embedded in physical machines or in software, and the promising capacities of both architectures makes it necessary for society and politicians to regulate the functions and limits of these devices \cite{Stone}. Despite the myth of destructive AI represented in films such as \textit{Terminator} or \textit{I, Robot}, the truth is that nowadays most usual smart algorithms consist of a series of simple rules applied to large series of numbers, and the result of that is called Artificial Intelligence. 

British Parliament and other institutions recommend the education from high school of Artificial Intelligence \cite{BritishParliament}, regardless of the pace of development of this technology, in order to cope future technological and social challenges.

The main reason is to improve technological understanding, enabling people to navigate an increasingly digital world, and inform the debate around how AI should, and should not, be used.

Computational thinking is a problem-solving method that uses techniques typically practiced by computer-scientists and is increasingly being viewed as an important ingredient of STEM learning in primary and secondary education \cite{Eguchi2014}. On the other hand, several studies in higher education research report on the low student pass rate in mathematics. Therefore, research into mathematics education has been greatly emphasized in the last decade \cite{Divjak2007}. Though education reform efforts have been made around the world, the trouble lies on the fact that most of schools are trying to prepare students for the future by continuing what was done in the past \cite{Eguchi2014}.

Thus, in this paper an educational proposal for teaching the basic mathematics behind simple AI algorithms is presented. The solution is developed in an open source platform, Scratch, and the main objective for the students is to be aware of the rules behind intelligent systems rather than learning or memorizing anything. The specific tasks to understand are automatic clustering, learning and prediction with AI. The software is designed for students of 16-18 years. The algorithms chosen for teaching are specifically designed or adapted to the mathematical background of the students. Moreover, the article will be publicly available in a web repository\footnote{\url{https://gitlab.com/juleste}}.

The present article is divided in the following sections. In section \ref{sec:Background} a background for teaching AI is presented. In Section \ref{sec:Content} the mathematics that will be used in the software are described. In Section \ref{sec:Methodology} the methodology that teachers will use with the students is detailed. Finally, in Section {\ref{sec:Conclusions}}, conclusions of the experiment are analyzed.

\vspace{7pt}
\section{Background}\label{sec:Background}
There is a wide consensus among computer scientist that it is quite difficult to teach the basics of AI \cite{Hearst1995}. This is due to the lack of a unified methodology and to the blend of many other disciplines which involve a wide range of skills, ranging from very applied to quite formal \cite{Kumar1998}. AI modeling, algorithms and applications may be taught using tools as simple as paper and pencil, traditional computer programming or hands-on-computer programming \cite{Greenwald}. Several papers discuss the prerequisites needed to understand machine learning \cite{Lavesson2010}. 

Some experts suggest that high school education reforms should encompass a drive on STEM skills and coding in schools; others support the idea of focusing on digital understanding, rather than in skills.

Nowadays, however, the tendency is to teach coding to students, mostly framed in robotics workshops \cite{Eguchi2014, Barker2007, Petre2004}. Seymour Papert \cite{Papert1980} laid much of the groundwork for using robots in the classroom in the 1970s. An argument for teaching to children and students using robots is that they see these machines as toys \cite{Mauch2001}. Studies show that robotics generates a high degree of student interest and engagement and promotes interest in maths and science careers.

\cite{Kumar1998} uses a Lego robot to make the students build the hardware and control the software that manages the sensors and the environment. \cite{Greenwald} uses a low-cost robot platform to teach students how a neural network is trained and built for a navigation robotic problem. There is the possibility that students pay attention to the robotic exercise rather than to the AI learning fundamentals, and this is what \cite{Kumar2004} tries to settle down in another robotic navigation problem. Both authors specify which are the premises to be transmitted to the students about the objective of the activity. These premises are, first, underlining that the exercise is about Artificial Intelligence and not robotics; second, emphasizing that they are clearly defined open projects, with specific start and end points.

However, although robotics is recognized as a proper way for teaching computational-thinking (CT) skills to students \cite{Grover2013}, in this article we present a wider approach to AI, based on the teaching of some mathematics with the help of Scratch \cite{Resnick2009}, which will let us introduce the students into more than one algorithm. 

CT can be defined as the process of recognizing aspects of computation in the world that surrounds us and applying tools and techniques from Computer Science to understand and reason about both natural and artificial systems and processes \cite{Grover2013,RS2012}. 

The software used to enhance CT on students should be easy for a beginner to overcome the difficulty of creating working programs, but it should also be powerful and extensive enough to satisfy the needs of more advanced programmers. Several programming tools fit these criteria in varying degrees: Scratch, Crickets, Karel the Robot, Alice, Game Maker, Kodu, Greenfoot and Agentcubes. All those tools are based on the Logo philosophy \cite{Papert1980}. Graphical programming environments are relatively easy to use and allow early experiences to focus on designing and creating, avoiding issues of programming syntax.

From a pedagogical perspective, computational tools are capable of deepening the learning of mathematics and science contents \cite{Guzdial1994,Eisenberg2002}, and the reverse is also true \cite{Weintrop2016}. Scratch is today one of the most popular of these programming environments, and it has proved to be very effective for the engagement and motivation of young students with no experience at programming \cite{Malan2007}.

\vspace{7pt}
\section{Content}\label{sec:Content}

This article will depict algorithms that can be practiced using Scratch in a workshop with students 16-18 years old. After the workshop, the students should be able to understand, play and eventually code these algorithms.

Specific algorithms are:
\begin{itemize}
\item K-means
\item Neural network
\end{itemize}

And the exercises that students will have to do are described next.

\textbf{K-means}

The algorithm K-means, developed in 1967 by MacQueen \cite{MacQueen}, is one of the simplest unsupervised learning algorithms that solve the well known clustering problem. K-means is a clustering algorithm which tries to show the students how items in a big dataset are automatically classified, even while new items are being added. The adopted technique for creating as many clusters as we want, is the minimum square error rule.

\begin{figure}[!ht]
\begin{center}
\setlength\fboxsep{0pt}
\setlength\fboxrule{0.5pt}$
\begin{array}{c}
\fbox{\includegraphics[width=7cm]{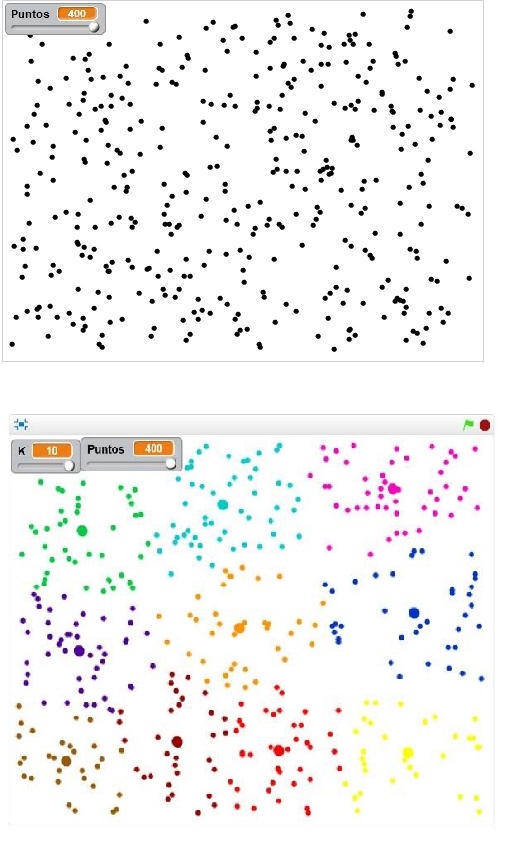}} \\

\end{array}$
\end{center}
\caption{Initial and final clouds of points}
\label{Kmeanscloud}
\end{figure}

Starting from a given cloud of N points and a smaller cloud of K mass centers, the aim of this activity is that the student learns how to cluster the points into K groups, by programming an application in Scratch. In order to do the clustering, each point will belong to the cluster defined by the closest mass center. Finally, each of the K clusters will be colored with a different color (see Figure \ref{Kmeanscloud}).

\textbf{Neural network}

The basic idea behind a neural network is to simulate lots of densely interconnected brain cells inside a computer so you can get it to learn things, recognize patterns, and make decisions in a human-like way. The main characteristic of this tool is that a neural network learns all by itself. The programmer just needs to design the physical structure (number of outputs, inputs, hidden layers) and set some very simple rules involving additions, multiplications and derivatives. They are based on perceptrons, which were developed in the 1950s and 1960s by the scientist Frank Rosenblatt \cite{Rosenblatt}, inspired by earlier work by Warren McCulloch and Walter Pitts \cite{McCulloch}. It is important to note that neural networks are (generally) software simulations: they are made by programming very ordinary computers. Computer simulations are just collections of algebraic variables and mathematical equations linking.

Habitually, neural networks use backpropagation type algorithms which require the usage of derivatives. However, the target students of this exercise (16-18 years old) do not have this mathematical operation in their curricula yet. As a consequence, the neural networks have been created with a logic activation function, which are understandable by the students. Necessary formulas are presented to them without any previous mathematical demonstration.

\begin{figure}[!ht]
\begin{center}
\setlength\fboxsep{0pt}
\setlength\fboxrule{0.5pt}
\fbox{\includegraphics[width=8cm]{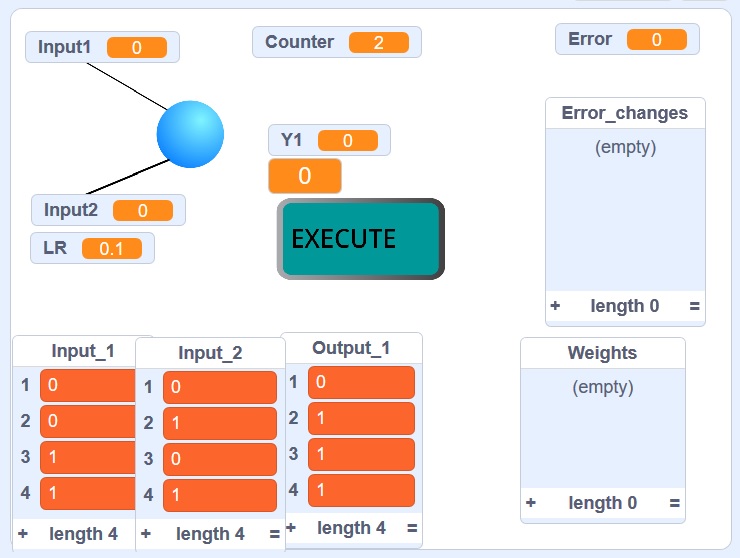}}
\caption{First neural network interface}
\label{FirstNNinterface}
\end{center}
\end{figure}

Different exercises are developed with neural networks. First, a simple neural network with two inputs and an output neuron, is trained with an AND logic gate (see Figure \ref{FirstNNinterface}). In each iteration, students will see the different weights that the neural network gets. Next, and OR logic gate will be used, and as a consequence, students will observe how adjusting parameters change with this new data. The mathematical description of this neural network is described in Subsection \ref{subsec:simpleNN}.

\begin{figure}[!ht]
\begin{center}
\setlength\fboxsep{0pt}
\setlength\fboxrule{0.5pt}
\fbox{\includegraphics[width=6cm]{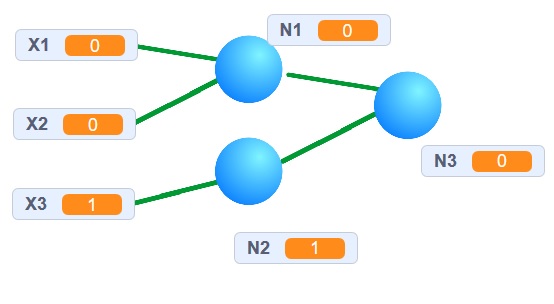}}
\caption{Second neural network interface}
\label{fig:SecondtNNinterface}
\end{center}
\end{figure}

As a second exercise of this type, a more complex neural network exercise is presented (Figure \ref{fig:SecondtNNinterface}. Considering a 3-2-1 neural network (three inputs, a hidden layer with two neurons and an output) the students will train the neural network with AND and OR logic gates again. The detailed operations of this exercise appear in Subsection \ref{subsec:complexNN}.

\subsubsection{Simple Neural Network operation}\label{subsec:simpleNN}

The neuron obtains an output $Y_1$ from the two inputs $Input_1$ and $Input_2$ using the corresponding weights.

\begin{equation}
\label{eq:erroreq}
N_1 = Input_1 \cdot W_1 + Input_2  \cdot W_2
\end{equation}

The activation function at $N_1$ defined as shown next:
\begin{equation}
    N_1 \begin{cases} 
    \geq 1 &\text{$Y_1$=1}\\\\
    \text{else}   &\text{$Y_1$=0}
    \end{cases}
\end{equation}

where $Y_1$ is the output of the neural network.

The error is defined as the difference between the desired output $Desired\_output$ and the obtained output.

\begin{equation}
\label{eq:erroreq}
Error = Desired\_output - Y_1
\end{equation}

Each time the function is executed, the algorithm updates the weights using backpropagation and gradient descent rule, until output $Y_1$ converges to the desired output $Desired\_output$.

The new value of the first weight will be the sum of its previous value and the product of the first input, the learning rate (LR) and the error.

\begin{equation}
\label{eq:W1eq}
W_{1,new} = W_1 + Input_1 \cdot LR \cdot Error
\end{equation}

Similarly, the new value of the second weight will be:

\begin{equation}
\label{eq:W2eq}
W_{2,new} = W_2 + Input_2 \cdot LR \cdot Error
\end{equation}

\subsubsection{Complex Neural Network operation}\label{subsec:complexNN}

There are three neurons: two at the hidden layer ($N_1$ and $N_2$) and one at the output ($N_3$). Their calculus depend on the inputs ($X_1$, $X_2$ and $X_3$) and weights ($W_1$, $W_2$, $W_3$, $W_4$, $W_5$) following next expressions: 

\begin{equation} \label{eq:N1eq}
\begin{cases}
N_1=X_1 \cdot W_1+ X_2 \cdot W_2 \\
N_2=X_3 \cdot W_3\\
N_3=N_1 \cdot W_4 + N_2 \cdot W_5
\end{cases}
\end{equation}

There is an activation function at $N_3$ defined as shown next:
\begin{equation}
    N_3 \begin{cases} 
    \geq 0 &\text{$N_3=1$}\\\\
    \text{else}   &\text{$N_3=0$}
    \end{cases}
\end{equation}
where $N_3$ is the output of the neural network.

Error is defined as:

\begin{equation}
\label{eq:erroreq}
Error = Desired\_output - N_3
\end{equation}

Employing again the backpropagation and gradient descent rule, the calculus for updating the weights result in next equations:

\begin{equation}
\begin{cases}
W_{1,new}=W_1 + LR \cdot Error \cdot W_4 \cdot X_1 \\
W_{2,new}=W_2 + LR \cdot Error \cdot W_4 \cdot X_2 \\
W_{3,new}=W_3 + LR \cdot Error \cdot W_5 \cdot X_3 \\
W_{4,new}=W_4 + LR \cdot Error \cdot N_1\\
W_{5,new}=W_5 + LR \cdot Error \cdot N_2 
\end{cases}
\end{equation}

Once the values of the error and weights are calculated, the student has to store them on the corresponding neurons.

\section{Methodology}\label{sec:Methodology}

The workshop of Artificial Intelligence will be based on the usage of the educational tool Scratch \cite{Resnick2009}. Scratch is a visual programming language, and its online community is targeted primarily at children and young students. Using Scratch, users can create online projects and develop them into almost anything by using a simple block-like interface. When they are finished, students can share their projects and discuss their creations with each other. Before the students start the exercises here proposed, it is convenient that they acquire some knowledge of this language.

In all algorithms presented in Section \ref{sec:Content}, the first task of the student is to fill in the white gaps left to him/her among the lines of code, marked with a comment. That is, the students do not need to create the algorithm or write the whole code themselves; the code will be provided for the most part.

Students will work in pairs and the workshop will be structured in the following steps:

\begin{itemize}

\item At the beginning, the teachers will give a short explanation of 15-20 minutes about AI and the objective of the workshop, with a twofold aim: first, to demystify Artificial Intelligence; second, to understand some simple mathematics underneath the computations.

\item The students will work in couples. Teachers will provide them with some written theoretical background about the algorithms and the instructions for the exercises (K-means and neural networks). Moreover, teachers will explain in another 10 minutes of presentation the basics of the algorithms.

\item Students will have one hour to finish the codes (20 minutes for K-means and 40 minutes for both parts of the neural networks) and eventually execute the applications to see whether they work properly. During this personal tasks teachers will be at hand ready to assist whenever necessary.

\item Finally, at the end of the session, the teachers will provide the students with the finished proposed answers in Scratch software for the students to check them.

\end{itemize}

\subsection{K-Means}

The algorithm is developed using one main block and three function blocks of Scratch code. The function blocks are called \texttt{NewDataSet}, \texttt{KMeans} and \texttt{ColourPoints}.

The student must finish up the three function blocks: the block \texttt{NewDataSet}, block \texttt{KMeans} and block \texttt{ColourPoints}. 

\begin{figure}[!ht]
\begin{center}
\setlength\fboxsep{0pt}
\setlength\fboxrule{0.5pt}
\fbox{\includegraphics[width=8cm]{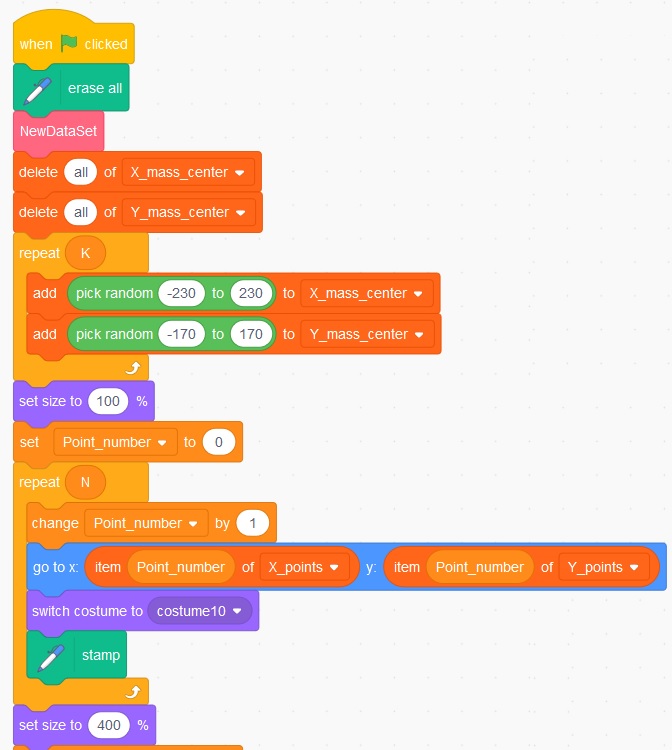}}
\caption{Main block to create the mass centers and the cloud of points}
\label{FunctionMassCenterCreation}
\end{center}
\end{figure}

The main block will create the mass centers and the cloud of points. The set of mass centers contains $K$ random points with $X$ coordinates between $\left( -230,230 \right)$ and $Y$ coordinates between $\left( -170,170 \right)$. These $K$ points will be the mass centers of the clusters (see Figure \ref{FunctionMassCenterCreation}).

\begin{figure}[!ht]
\begin{center}
\setlength\fboxsep{0pt}
\setlength\fboxrule{0.5pt}
\fbox{\includegraphics[width=8cm]{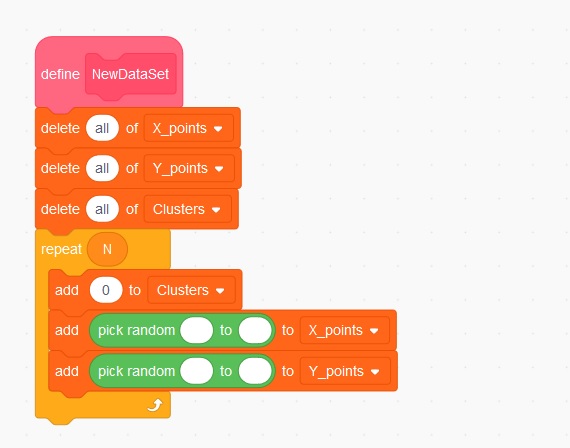}}
\caption{Task 1. Unfinished function to create the cloud of points}
\label{FunctionPointCreation}
\end{center}
\end{figure}

The first task is to finish the programming of the block \texttt{NewDataSet} that will create the cloud of $N$ points. The cloud must contain $N$ points with $X$ coordinates between $\left( -230,230 \right)$ and $Y$ coordinates between $\left( -170,170 \right)$ (see Figure \ref{FunctionPointCreation}).

Block \texttt{KMeans} stores the number of the mass center that will be assigned to each point. To do that, the student must code the calculation of the Euclidean distance from each of the points to each the mass centers in variable $A$. The program will then find the minimum distance of them all and fill up vector \texttt{Clusters}, which contains the numbers of the cluster to which each point belongs.

Finally, block \texttt{ColourPoints} graphs the clouds of points and the cloud of centers of mass, each in a colour given by vector \texttt{Clusters}.

The algorithm here presented just tries to show automatic clustering, and not getting equivalent size clusters.

\begin{figure}[!ht]
\begin{center}
\setlength\fboxsep{0pt}
\setlength\fboxrule{0.5pt}

\fbox{\includegraphics[width=8cm]{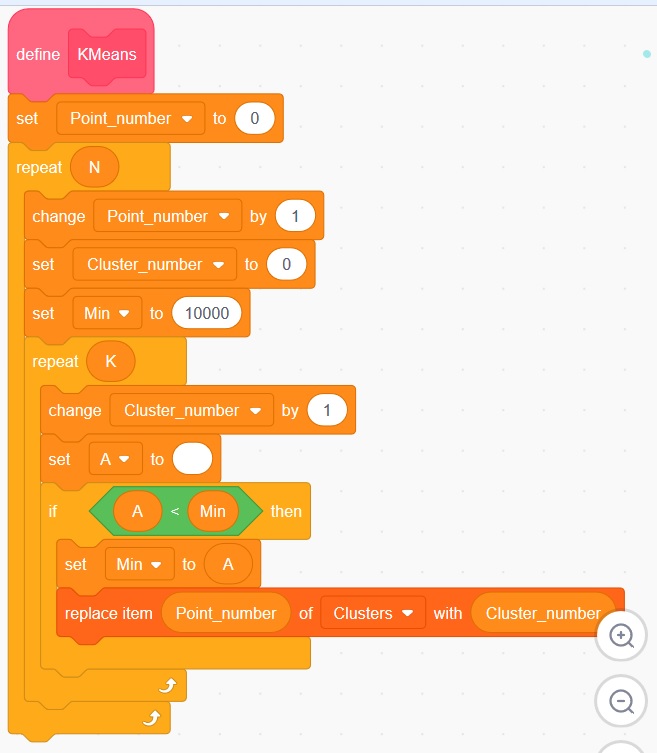}}
\caption{Task 2. Unfinished function to calculate the distances $A$}
\label{Avariable}
\end{center}
\end{figure}

\subsection{Neural Network: AND/OR logic gate}
The algorithm is developed using a main block, which initializes the data, and two blocks, \texttt{Neuron} and \texttt{ExecuteButton}.

\begin{figure}[!ht]
\begin{center}
\setlength\fboxsep{0pt}
\setlength\fboxrule{0.5pt}
\fbox{\includegraphics[width=8cm]{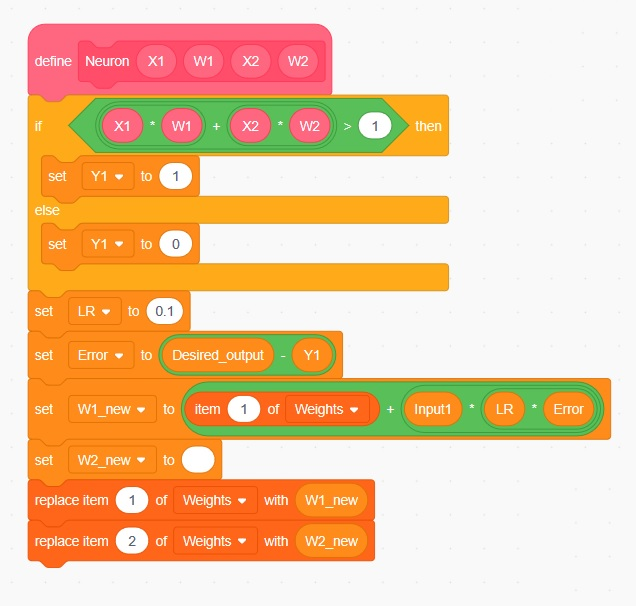}}

\caption{Task 3. Unfinished function to calculate the weight update}
\label{fig:SimpleNeuron}
\end{center}
\end{figure}

The interface used is described in Figure \ref{FirstNNinterface}. The student must code the equations that define the function that performs the only neuron of the network. The equations to calculate is the update of $W_2$, as can be seen in Figure \ref{fig:SimpleNeuron}. 

The students must train the neuron using two sets of data: one set for the AND logic gate and another for the OR logic gate.

\subsection{Complex Neural Network Operation}
This exercise is based on training a three inputs neural network with AND and OR logic gates. As an additional complexity comparing to previous exercise, this neural network is multilayered. It uses three neurons, \texttt{Neuron1}, \texttt{Neuron2} and \texttt{Neuron3}, and \texttt{ExecuteButton}.

The exercise is planned to fill the gaps of the following operations:
\begin{itemize}
    \item Net calculus of $N_3$
    \item Update of $W_2$ and $W_5$
    \item Update of $B_2$
\end{itemize}

\vspace{7pt}

\vspace{7pt}
\section{Conclusions}\label{sec:Conclusions}
The paper presents teacher-guided, easy to implant activities that can be performed at schools using Scratch. Moreover, the operations have been adapted for 16-18 year-old students' mathematical background.

The tasks presented are scalable; students can delve into the maths involved in the mathematical iterations or into the Scratch code itself, or even propose new neural networks to deal with other problems. 

The work can be extensible to students of different ages and more AI algorithms can be added to the system. 

The easiness of the equations of K-means and neural networks permit their implementation in other formats, such as MS Excel, which in some occasions could be more familiar to students than Scratch.

Moreover, the students will realize that the mathematical knowledge acquired throughout the year will help them finishing the programming of automatic clustering and neuron networks, being able to train AND and OR logic gates.

As a future work, first, the authors should organize the AI workshop several times and measure the grade of the objective achievement in collaboration with pedagogical researchers. Secondly, more programming languages should be explored (GeoGebra, HTML, ShynnyApps) in order to implement more exercises, such as the usage of neural networks for data prediction.


\vspace{7pt}

\vspace{7pt}

\section{Acknowledgements}
This project has received funding from the European Union's Horizon 2020 research and innovation programme under the Marie Sklodowska-Curie grant agreement No 777720.


\vspace{7pt}
\begin{thebibliography}{1}

\bibitem{Arimoto}
Tateo Arimoto and Yasushi Sato. Rebuilding public trust in science for policy-making. Science, 337(6099):1176-1177, 2012.

\bibitem{Haerlin}
Benny Haerlin and Doug Parr. How to restore public trust in science. Nature,
400(6744):499, 1999.

\bibitem{MacQueen}
MacQueen. Some methods for classification and analysis of multivariate observations. In Proceedings of the Fifth Berkeley Symposium on Mathematical Statistics and Probability, Volume 1: Statistics, pages 281-297, Berkeley, Calif., 1967. University of California Press.

\bibitem{McCulloch}
Warren S McCulloch and Walter Pitts. A logical calculus of the ideas immanent in nervous activity. The bulletin of mathematical biophysics, 5(4):115-133, 1943.

\bibitem{Nielsen}
Michael A Nielsen. Neural networks and deep learning. Determination Press, 2015.

\bibitem{OECD}
OECD. Scientific advice for policy making: The role and responsibility of expert bodies and individual scientists. Technical Report 21, OECD Science, Technology and Industry Policy Papers, 2015.

\bibitem{Rosenblatt}
Rosenblatt, F. (1958). The perceptron: a probabilistic model for information storage and organization in the brain. Psychological review, 65(6), 386.

\bibitem{Stone}
Peter Stone, Rodney Brooks, Erik Brynjolfsson, Ryan Calo, Oren Etzioni, Greg Hager, Julia Hirschberg, Shivaram Kalyanakrishnan, Ece Kamar, Sarit Kraus, et al. Artificial intelligence and life in 2030. One Hundred Year Study on Artificial Intelligence: Report of the 2015-2016 Study Panel, 2016.

\bibitem{Woodford}
C Woodford. How neural networks work - a simple introduction, 2016.

\bibitem{Greenwald}
Greenwald, L., Artz, D. (2004). Teaching artificial intelligence with low-cost robots. Accessible hands-on artificial intelligence and robotics education, ed. L. Greenwald, Z. Dodds, A. Howard, S. Tejada, and J. Weinberg, 35-41.

\bibitem{Hearst1995}
Hearst, M. (1995). Improving Instruction of Introductory Artificial Intelligence: Papers from the 1994 AAAI Fall Symposium, November 4-6, New Orleans, Louisiana. AAAI Press.


\bibitem{Kumar1998}
Kumar, D., Meeden, L. (1998). A robot laboratory for teaching artificial intelligence. ACM SIGCSE Bulletin, 30(1), 341-344.

\bibitem{Eguchi2014}
Eguchi, A. (2014, July). Robotics as a learning tool for educational transformation. In Proceeding of 4th International Workshop Teaching Robotics, Teaching with Robotics, 5th International Conference Robotics in Education Padova (Italy).

\bibitem{Papert1980}
Papert, S. (1980). Mindstorms: Children, computers, and powerful ideas. New York:  Basic Books, Inc.

\bibitem{Kumar2004}
Kumar, A. N. (2004). Three years of using robots in an artificial intelligence course: lessons learned. Journal on Educational Resources in Computing (JERIC), 4(3), 2.

\bibitem{Barker2007}
Barker, B. S.,  Ansorge, J. (2007). Robotics as means to increase achievement scores in an informal learning environment. Journal of research on technology in education, 39(3), 229-243.

\bibitem{Petre2004}
Petre, M., Price, B. (2004). Using robotics to motivate ‘back door’ learning. Education and information technologies, 9(2), 147-158.

\bibitem{Mauch2001}
Mauch, E. (2001). Using technological innovation to improve the problem-solving skills of middle school students. 7he Clearing House, 75(4), 211-213.

\bibitem{Lavesson2010}
Lavesson, N. (2010). Learning machine learning: a case study. IEEE Transactions on Education, 53(4), 672-676.

\bibitem{Divjak2007}
Divjak, B., Erjavec, Z. (2008). Enhancing Mathematics for Informatics and its correlation with student pass rates. International journal of mathematical education in science and technology, 39(1), 23-33.

\bibitem{Resnick2009}
Resnick, M., Maloney, J., Monroy-Hernández, A., Rusk, N., Eastmond, E., Brennan, K., Kafai, Y. (2009). Scratch: programming for all. Communications of the ACM, 52(11), 60-67.

\bibitem{Grover2013}
Grover, S., Pea, R. (2013). Computational thinking in K–12: A review of the state of the field. Educational Researcher, 42(1), 38-43.

\bibitem{RS2012}
Royal Society. (2012). Shut down or restart: The way forward for computing in UK schools. Retrieved from http://royalsociety.org/education/policy/computing-in-schools/report/.

\bibitem{Guzdial1994}
Guzdial M (1994) Software-realized scaffolding to facilitate programming for science learning. Interact Learn Environ 4(1):001-044.

\bibitem{Eisenberg2002}
Eisenberg M (2002) Output devices, computation, and the future of mathematical crafts. Int J Comput Math Learn 7(1):1-44.

\bibitem{Weintrop2016}
Weintrop, D., Beheshti, E., Horn, M., Orton, K., Jona, K., Trouille, L., Wilensky, U. (2016). Defining computational thinking for mathematics and science classrooms. Journal of Science Education and Technology, 25(1), 127-147.

\bibitem{Malan2007}
Malan, D. J., Leitner, H. H. (2007, March). Scratch for budding computer scientists. In ACM Sigcse Bulletin (Vol. 39, No. 1, pp. 223-227). ACM.

\bibitem{BritishParliament}
AI in the UK: ready, willing and able? - Tech report (2018). Select Committee on Artificial Intelligence. House of Lords.

\end{thebibliography}
\end{document}